# PROFESSIONAL CERTIFICATION BENCHMARK DATASET: THE FIRST 500 JOBS FOR LARGE LANGUAGE MODELS

David Noever and Matt Ciolino
PeopleTec, Inc., Huntsville, AL, USA
David.noever@peopletec.com    matt.ciolino@peopletec.com

## ABSTRACT

*The research creates a professional certification survey to test large language models and evaluate their employable skills. It compares the performance of two AI models, GPT-3 and Turbo-GPT3.5, on a benchmark dataset of 1149 professional certifications, emphasizing vocational readiness rather than academic performance. GPT-3 achieved a passing score (>70% correct) in 39% of the professional certifications without fine-tuning or exam preparation. The models demonstrated qualifications in various computer-related fields, such as cloud and virtualization, business analytics, cybersecurity, network setup and repair, and data analytics. Turbo-GPT3.5 scored 100% on the valuable Offensive Security Certified Professional (OSCP) exam. The models also displayed competence in other professional domains, including nursing, licensed counseling, pharmacy, and teaching. Turbo-GPT3.5 passed the Financial Industry Regulatory Authority (FINRA) Series 6 exam with a 70% grade without preparation. Interestingly, Turbo-GPT3.5 performed well on customer service tasks, suggesting potential applications in human augmentation for chatbots in call centers and routine advice services. The models also score well on sensory and experience-based tests such as wine sommelier, beer taster, emotional quotient, and body language reader. The OpenAI model improvement from Babbage to Turbo resulted in a median 60% better-graded performance in less than a few years. This progress suggests that focusing on the latest model's shortcomings could lead to a highly performant AI capable of mastering the most demanding professional certifications. We open-source the benchmark to expand the range of testable professional skills as the models improve or gain emergent capabilities.*

## Keywords

## 1. INTRODUCTION

There is growing interest in novel ways to score AI progress using large language models (LLMs) and their remarkable ability to pass challenging professional exams [1-47]. These certifications [4-5, 19, 40] include fields of medicine [6-7,15,17,20,23-25,27-28,30-31,36], law [13], physics [43], math [16,41], engineering [34], software [14], psychology [32], IQ [39], finance [37], language translation [8,10-12,21,29,44], and general task solving [1-2,9,18,26,35,38,42]. A new field seeks to design cheat-proof examination styles that neutralize a student's attempt to use an LLM while testing [40]. The original Turing test [48-49] posed a conversational test if a machine could convince a blind human panel that the judges conversed with a human when they talked to software only. Researchers have scored LLMs grammatically or, in the case of translators using the Bilingual Evaluation (BLEU) methodology to compare performance to experts and other machines [8,10-12,21,29,44]. However, with the remarkable progress in language understanding since late 2022 (ChatGPT and other transformers [3-4]), the current LLM generation has received attention for its capability to certify as human professionals in often complex or even life-saving capacities like legal advisors [13] or medical internists [6]. Remarkably, early prototype systems feature suicide prevention and mental health counseling as the first applications of LLMs, given some safeguards [50]. The increasing importance of understanding the result of training, refining, and deploying these models has prompted OpenAI [4] to publish its "eval corpus," [51] mainly a description of the internal stages of declaring a model safe to deploy.

The goal of the current work stems from a motivational thought experiment: What would an employer do if presented with ChatGPT's resume in a job interview? In other words, is there a role to consider the machine as both a trustworthy personal assistant and as a prospective new hire? [22] The question becomes more interesting when considering the model's previous accomplishments in taking difficult exams [4]. ChatGPT (GPT 4.0) has passed the medical [6] and legal bar [14] exams. It certifies as a software programming competitor in Code Forces [4]. As a math and biology Olympiad winner, GPT4 passes all Advanced Placement exams and ranks in the top 95% for college admission (Scholastic Aptitude Test) [4]. In this context, a machine looking for a job carries an initial credential worthy of full college admission for undergraduate and potentially graduate schoolwork [4,42].

Could the machine show similar capabilities as a professional employee or job candidate in the applicant pool with human experts, particularly considering the $20 per month charge for priority use of the OpenAI API? A few hundred dollars for a tireless personal assistant seem at face value to be a productive use of funds.

The current research presents a new benchmark [52] called the Professional Certification Benchmark Exam Dataset, based on a representative sampling of more than 1100 practice exams for professional certifications (5197 questions) [53]. In addition, the benchmark includes synthetically constructed examples of 49 challenging skill tests highlighted from human surveys as challenging [54]. The synthetic dataset uses proposed 10-20 question panels and a separate session to answer the question (940 total). This new assessment or total panel comprises 1149 occupational skill evaluations ranging from high sensory qualifications for a wine sommelier to rigorous offensive cybersecurity tests. The new dataset [52] consists of 1149 exams with 6137 multiple questions for the model to answer.

The work builds on previous Open AI internal assessments of large language models [4], mainly focusing on the medical and legal profession (bar exams) and various scholastic achievement or advanced placement exams. One goal of these assessments ranks the improvement in general "zero-shot" knowledge [18] available as the large language model (LLM) progresses in sophistication and capability. In this context, zero-shot refers to an LLM not requiring any example cases. Few-shot direct prompt engineering compares to a lazy human candidate with some study preparations.

## 2. METHODS

The research approach centers on evaluating LLMs by surveying these large banks of practice tests, including all the certifications that typically rely on multiple-choice selections. In this way, we score the progress over the last two years in building confident exam takers for various professional jobs, ranging from wine sommeliers to certified accountants. This work looks at 1149 such certifications available in the public domain as practice tests (mainly) but provides a wide swath of specializations. The approach attempts two styles of inquiry. Firstly, we survey all 1100 practice exams of professions [53], then the 49 certifications in a shallow fashion to illustrate the range of knowledge. Secondly, we finish the equivalent of a cybersecurity cloud certification that otherwise requires an applicant to study a recommended six-month preparation before examination. In addition to this variation of test types, we explore several generations of model complexity, including three or more from the OpenAI line-up of GPT, then branching to open-source alternatives. All test cases are multiple-choice, the preferred standard for most certifications to assist automated and objective grading methods. We evaluate the results against at least four OpenAI models released as API interfaces since 2021. In recency and performance, these models follow alphabetically famous innovators, including Ada (Lovelace), (Charles) Babbage, (Marie) Curie, and (Leonardo DaVinci). DaVinci is the best available model (2022) until the upgrade of GPT3 to GPT4, which OpenAI has not included in API formats (May 2023). The latest performant model is labeled gpt3.5-turbo.

Collected metrics include model type, the score of correct multiple-choice questions, and models that pass the professional certification exam (>70% accurate). All models are tested as "zero-shot" learners, meaning that we prefix no specific examples of answering multiple-choice questions to the final prompt. In our experience, the "few shot" method of example proved unnecessary, as all the models tested followed instructions to choose a letter answer (A-D) representing the model's response. For all models tested, the prefix instruction included the statement, "You are an expert in professional certifications in <<BLANK>>. Answer the multiple-choice questions using the format Answer: [insert]," where <<BLANK>> represented the broad category of the exam such as "NIST SP800-171 CMMC Certification". For models where the API exposes the creativity level or temperature, we set the value to 0 to limit the model drift in ways consistent with other OpenAI applications that center on a question and answer rather than creative outputs.

## 3. RESULTS

Table 1 summarizes the collected results for five example professional certifications administered to

| Table 1. Example Scores Based on Five Professional Examinations Administered to Two Models | | |
|---|---|---|
| Certification | Turbo-GPT3.5 | GPT3 |
| ALIBABA ACA CLOUD1 | 100% | 100% |
| COMPTIA 220 1102 | 100% | 100% |
| COMPTIA 220 902 | 100% | 100% |
| CompTIA XK0 005 | 100% | 100% |
| CSA CCSK | 100% | 100% |

two models, GPT3 and Turbo-GPT3.5. The Appendix lists the complete scoresheet for all 1149 professional certifications. Figure 1 highlights the overall difference between the two models and the rank order performance for both models that achieve a passing grade (>70% correct) using the zero-shot approach. GPT passed roughly 39% of the broad professional certifications without fine-tuning or exam preparation. The examinations covered a predominantly weighted selection of computer-related vocations, ranging from high-value cybersecurity certifications to project management. In contrast to previous work that selected exams based on academic performance, this benchmark dataset highlights vocational readiness. More than 100 cyber certifications include a recommended (human) employee preparation time of at least six months. In the case of offensive cybersecurity or red team, the Offensive Security Certified Professional (OSCP) certification is a valuable job skill for penetration testers, one that industry and government recruiters reward with substantial salary boosts. GPT3.5 scores 100% on the OSCP exam.

The test benchmark [52] covers many of the academic requirements to be considered professionally employable, ranging from the high school graduation exam (General Education Development (GED)) to graduate admission (Graduate Management Admission Test (GMAT)). Another noteworthy category addressed here and in previous testing [4], the model shows competent and specific skills in nursing (Test of Essential Academic Skills, or TEAS), licensed counselor (National Counselor Examination (NCE)) and pharmacy (North American Pharmacist Licensure Examination, NAPLEX). The model passes the teacher certification (PRAXIS), an essential requirement to participate in US public education employment. Other than the medical or legal bar exam, the Financial Industry Regulatory Authority (FINRA) Series 6 exam provides a lucrative gateway to financial advisers, one which has a human pass rate of 58% [55] and one which GPT-3.5 scores a passing (70%) grade without preparation.

Based on tabulated professional certification results (Appendix), the latest LLMs demonstrate qualifications in broad computer-related fields, including cloud and virtualization (Amazon AWS, Alibaba, Google, Microsoft, VMWare), business analytics (Project Management Institute, Six Sigma, Configuration Management), cybersecurity (CompTIA, EC Council, InfoSec Institute, IAPP, Solarwinds, Palo Alto Networks, Blockchain, etc.), network setup and repair (Dell, HP, Citrix, Novell, Fortinet), data analytics (Tableau, Qlikview Developer, UIPath).

Ironically, a chat interface (turbo-gpt3.5) scores well on traditional customer service responsibilities tested by Avaya exams, a likely early destination for human augmentation with chatbots for routine advice and call centers.

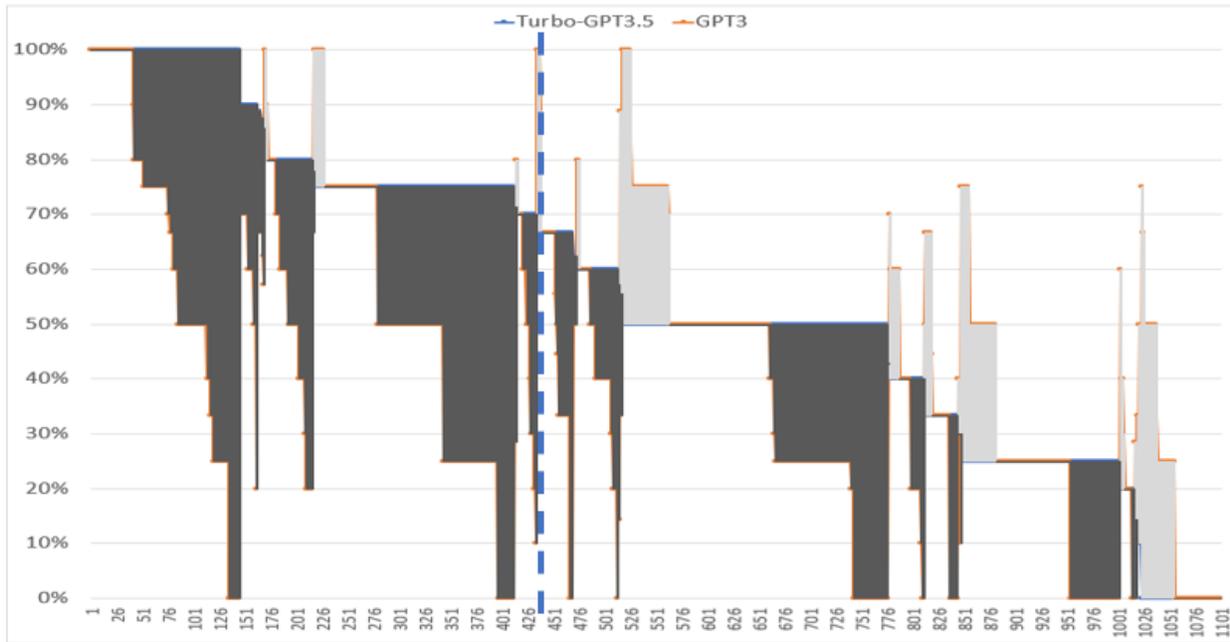

*Figure 1. Rank ordered professional exam scores between two LLM models, rated by the percentage of questions answered correctly. The blue dotted line shows the 435 examinations scoring 70% or higher with zero-shot approach.*

Notable certification failures overlap with some of the successful passing areas, such as VMWare (virtualization), teaching writing (Praxis), and logical legal reasoning (LSAT Section 1, Logical Reasoning). Further work is needed to understand whether these multiple-choice exams fit the correct format for prompting the LLMs and whether the test evaluation proves appropriate for automated scoring. For example, the low scores associated with writing and Python coding point to some prompt ambiguity in the question submission since LLMs have demonstrated high qualifications for grammar, language comprehension, and coding [4].

To evaluate the models' behavior, Figure 2 highlights challenging but non-computer-related tasks like selecting wine (sommelier exams), language translation, driver's education, and various IQ tests (Mensa, Wonderlic). One motivation for this list was their high rank as the most challenging certifications [54]. Examples here included accountants (CPA), veterinarians (CVPM), aviation inspectors (CMI-ASQ, CAM), real estate appraisers, human resources professionals (SHRM), and financial planners (CFP).

Another list motivator hinged on finding exams that relied on sensory or emotional experience, which in the case of LLMs relies on acquired knowledge, not direct experience. Examples include wine and beer tasting, emotional intelligence assessment (EQ), body language interpretation, and driver's education.

Figure 2 shows that turbo-gpt3.5 passes all exams in this group except for Wonderlic (50%) with greater than 70% passing grades. The average human score on Wonderlic is 40% (20/50 correct). No noticeable bias appears in the experience-based tests like sensory or emotional evaluations, as these text generation examples exist in the training set of many internet pages used to build and fine-tune the model responses. The notable improvement in the OpenAI model improvement from Babbage to Turbo approaches a median 60% better-graded performance (100% in Turbo vs. 40% in Babbage). This advance has taken less than a few years and suggests that with sufficient attention to the latest model's failings in Appendix examples, a highly performant model could master the most demanding professional certifications.

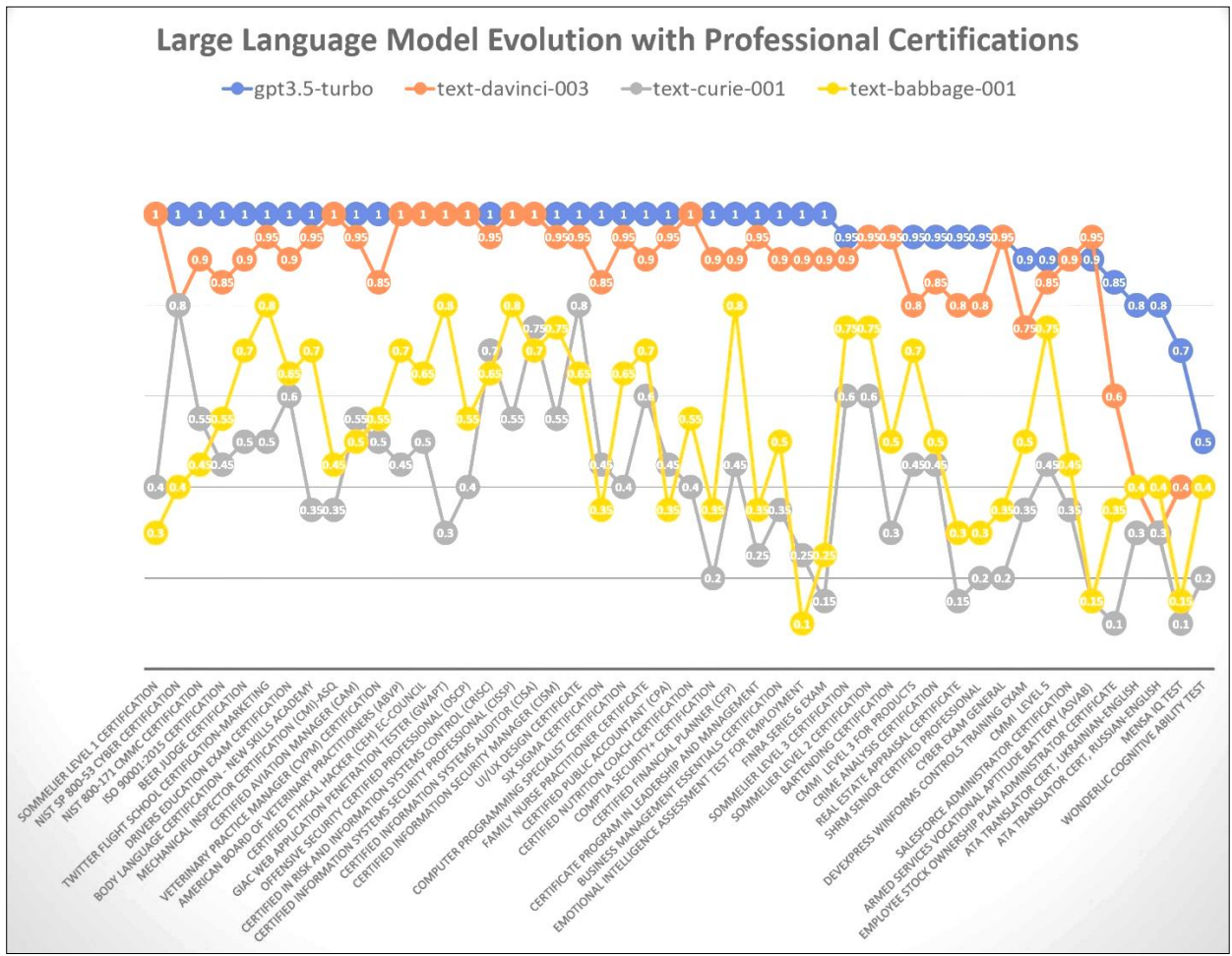

*Figure 2. Comparison of hardest certifications across different model evolutions, including experience-based tests*

## 4. DISCUSSION AND CONCLUSIONS

Three often mentioned AI axes include reliability, competencies, and trustworthiness outside its narrow training set. These trade-offs also trigger concerns about creating a large language model that lacks any verifiable model of the world, whether that model involves physics or common sense, or empathy. One outcome of recent attempts to engineer prompts that change the model's behavior creates concerns that the model serves precisely the role that the user requires, whether nefarious, malicious, or benign. The right prompt generates the proper response. Previous work has labeled these categories of AI safety as constraints on reasoning ability, hallucination, and user interactivity.

This study assembled a comprehensive professional certification benchmark and scored multiple language models to evaluate competency and capabilities without preparation or special attention to the examples for training. The latest models (turbo-gpt3.5) demonstrated competence in various medical, legal, technical, and financial areas of human expertise. The original contributions of the research created the benchmark dataset for future testing of new models against prior ones across a range of capabilities. The OpenAI API provides a scalable approach to expanding the list from 5197 questions to broader professional competencies.

During this phase of deserved attention to AI safety, these results probe the limits of AI trust and confidence. The work joins with similar quantitative evaluations provided by Open AI (called eval corpus [51]) and the recent studies illustrating how ChatGPT and GPT-4 models exceed (by 10-60%) the exam-taking tasks of its most recent predecessors across a spectrum of educational milestones [19]. Open AI presented a bar chart comparing a single GPT generation (3/4) and its progress in taking Advanced Placement (high school) exams, the legal and medical bar, the graduate GRE, the college SAT, and various competitive math and biology Olympiads [19]. Others [40] have criticized the testing industry for turning to many multiple-choice formats because of their ease of automated scoring, a feature that may assist the LLMs in guessing well when considering the next token prediction [2-3].

While these educational evaluations assess verbal, comprehension, and mathematical understanding, they focus on the core skills expected of general education. On the other hand, professional certifications provide confidence that a new hire possesses the prior training to exercise competence in specific fields. Areas of particular future interest involve more experience-based testing contributions, such that model parroting of training data becomes less influential in the final assessment. A notable feature of testing sensory and emotional qualifications reverses some widely held beliefs that this generation of AI would fail to interact naturally with humans as a personal assistant. On the contrary, these test results bear a high emotional quotient and some capability to engage in the impossible sensory evaluations in a quantitative conversational context.

## ACKNOWLEDGMENTS

The author would like to thank the PeopleTec Technical Fellows program for its encouragement and project assistance. The author thanks the researchers at Open AI for developing large language models and allowing public access to ChatGPT.

## Authors

**David Noever** has research experience with NASA and the Department of Defense in machine learning and data mining. He received his BS from Princeton University and his Ph.D. from Oxford University as a Rhodes Scholar in theoretical physics.

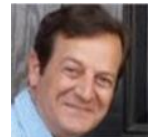

**Matt Ciolino** has research experience in deep learning and computer vision. He received his Bachelor's in Mechanical Engineering from Lehigh University. Matt is pursuing graduate study in computer vision and machine learning at Georgia Tech.

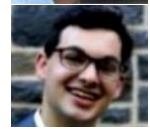

**Appendix A: Examination Evaluation Compared between Turbo GPT 3.5 vs. GPT 3**

| Certification | Turbo-GPT3.5 | GPT3 | Certification | Turbo-GPT3.5 | GPT3 |
|---|---|---|---|---|---|
| ALIBABA ACA CLOUD1 | 100% | 100% | AAFM INDIA CWM LEVEL 2 | 50% | 50% |
| COMPTIA 220 1102 | 100% | 100% | AMAZON AWS CERTIFIED DEVELOPER ASSOCIATE DVA C02 | 50% | 50% |
| COMPTIA 220 902 | 100% | 100% | AMAZON AWS CERTIFIED DEVOPS ENGINEER PROFESSIONAL DOP C02 | 50% | 50% |
| COMPTIA XK0 005 | 100% | 100% | AMAZON AWS CERTIFIED SOLUTIONS ARCHITECT ASSOCIATE SAA C03 | 50% | 50% |
| CSA CCSK | 100% | 100% | AMAZON AWS CERTIFIED SOLUTIONS ARCHITECT PROFESSIONAL | 50% | 50% |
| DELL DEA 2TT4 | 100% | 100% | ANDROIDATC AND 402 | 50% | 50% |
| DMI PDDM | 100% | 100% | ARUBA ACCP V62 | 50% | 50% |
| ECCOUNCIL 212 82 | 100% | 100% | AVAYA 3300 | 50% | 50% |
| EXIN MORF | 100% | 100% | AVAYA 3308 | 50% | 50% |
| GENESYS GE0 803 | 100% | 100% | AVAYA 6202 | 50% | 50% |
| GOOGLE CLOUD DIGITAL LEADER | 100% | 100% | AVAYA 7391X | 50% | 50% |
| GOOGLE GOOGLE ANALYTICS | 100% | 100% | BCS FCBA | 50% | 50% |
| GOOGLE PROFESSIONAL DATA ENGINEER | 100% | 100% | BCS ISEB SWT2 | 50% | 50% |
| IAPP CIPT | 100% | 100% | BCS RE18 | 50% | 50% |
| LPI 102 500 | 100% | 100% | CERTNEXUS ITS 110 | 50% | 50% |
| MICROSOFT 62 193 | 100% | 100% | CHECKPOINT 156 560 | 50% | 50% |
| MICROSOFT 70 334 | 100% | 100% | CITRIX 1Y0 403 | 50% | 50% |
| MICROSOFT 70 398 | 100% | 100% | COMPTIA PT1 002 | 50% | 50% |
| MICROSOFT 70 465 | 100% | 100% | CWNP CWDP 303 | 50% | 50% |
| MICROSOFT 70 475 | 100% | 100% | CYBERARK CAU201 | 50% | 50% |
| MICROSOFT 70 488 | 100% | 100% | DATABRICKS CERTIFIED DATA ENGINEER ASSOCIATE | 50% | 50% |
| MICROSOFT 70 489 | 100% | 100% | DELL DEA 1TT4 | 50% | 50% |
| MICROSOFT 98 349 | 100% | 100% | DELL DEA 1TT5 | 50% | 50% |
| MICROSOFT 98 361 | 100% | 100% | DELL DEA 41T1 | 50% | 50% |
| MICROSOFT 98 366 | 100% | 100% | DELL DEA 64T1 | 50% | 50% |
| MICROSOFT 98 382 | 100% | 100% | DELL DES 2T13 | 50% | 50% |
| MICROSOFT AI 102 | 100% | 100% | DELL DES 5221 | 50% | 50% |
| MICROSOFT AI 900 | 100% | 100% | DELL DES DD23 | 50% | 50% |
| MICROSOFT AZ 101 | 100% | 100% | DELL E20 368 | 50% | 50% |
| MICROSOFT AZ 140 | 100% | 100% | ECCOUNCIL 312 39 | 50% | 50% |
| MICROSOFT AZ 200 | 100% | 100% | ECCOUNCIL 312 50V12 | 50% | 50% |

| Certification | Turbo-GPT3.5 | GPT3 | Certification | Turbo-GPT3.5 | GPT3 |
|---|---|---|---|---|---|
| MICROSOFT AZ 801 | 100% | 100% | EXIN ASM | 50% | 50% |
| MICROSOFT MB 900 | 100% | 100% | FORTINET NSE4 FGT 60 | 50% | 50% |
| MICROSOFT MB 901 | 100% | 100% | FORTINET NSE5 EDR 5 0 | 50% | 50% |
| MICROSOFT MS 700 | 100% | 100% | FORTINET NSE5 FCT 7 0 | 50% | 50% |
| PMI PMI PBA | 100% | 100% | FORTINET NSE7 SDW 64 | 50% | 50% |
| SALESFORCE DEV 501 | 100% | 100% | GAQM CBAF 001 | 50% | 50% |
| SNIA S10 110 | 100% | 100% | GIAC GSNA | 50% | 50% |
| TEST PREP MCAT SECTION 1 VERBAL REASONING | 100% | 100% | GOOGLE GSUITE | 50% | 50% |
| TEST PREP TEAS SECTION 4 SENTENCE COMPLETION | 100% | 100% | GUIDANCE SOFTWARE GD0 110 | 50% | 50% |
| THE OPEN GROUP OG0 093 | 100% | 100% | HP HPE0 S51 | 50% | 50% |
| MICROSOFT 98 365 | 100% | 90% | HP HPE2 E67 | 50% | 50% |
| ANDROIDATC AND 401 | 100% | 80% | HP HPE2 T36 | 50% | 50% |
| COMPTIA PT0 002 | 100% | 80% | HUAWEI H12 811 | 50% | 50% |
| ECCOUNCIL 312 50V11 | 100% | 80% | IAPP CIPP A | 50% | 50% |
| GAQM CLSSGB | 100% | 80% | ISACA COBIT 2019 | 50% | 50% |
| IAPP CIPP E | 100% | 80% | ISQI CTAL TM SYLL2012 | 50% | 50% |
| ISC CISSP ISSAP | 100% | 80% | MAGENTO MAGENTO 2 CERTIFIED ASSOCIATE DEVELOPER | 50% | 50% |
| MICROSOFT 98 364 | 100% | 80% | MICROSOFT 70 332 | 50% | 50% |
| NACVA CVA | 100% | 80% | MICROSOFT 70 339 | 50% | 50% |
| NOKIA NOKIA 4A0 100 | 100% | 80% | MICROSOFT 70 342 | 50% | 50% |
| APICS CPIM BSP | 100% | 75% | MICROSOFT 70 348 | 50% | 50% |
| APPLE 9L0 012 | 100% | 75% | MICROSOFT 70 414 | 50% | 50% |
| ASQ CSSGB | 100% | 75% | MICROSOFT 70 487 | 50% | 50% |
| AVAYA 72200X | 100% | 75% | MICROSOFT 98 375 | 50% | 50% |
| EXIN ASF | 100% | 75% | MICROSOFT AI 100 | 50% | 50% |
| EXIN EX0 008 | 100% | 75% | MICROSOFT AZ 600 | 50% | 50% |
| EXIN EX0 115 | 100% | 75% | MICROSOFT DP 500 | 50% | 50% |
| FORTINET NSE5 FAZ 54 | 100% | 75% | MICROSOFT MB 330 | 50% | 50% |
| GAQM BPM 001 | 100% | 75% | MICROSOFT MB2 710 | 50% | 50% |
| GAQM CDCP 001 | 100% | 75% | MICROSOFT MB2 713 | 50% | 50% |
| GAQM CSM 001 | 100% | 75% | MICROSOFT MS 203 | 50% | 50% |
| GOOGLE PROFESSIONAL GOOGLE WORKSPACE ADMINISTRATOR | 100% | 75% | MICROSOFT MS 302 | 50% | 50% |
| IAAP CPACC | 100% | 75% | MICROSOFT PL 600 | 50% | 50% |
| IIBA IIBA AAC | 100% | 75% | NETAPP NS0 003 | 50% | 50% |

| Certification | Turbo-GPT3.5 | GPT3 | Certification | Turbo-GPT3.5 | GPT3 |
|---|---|---|---|---|---|
| ISACA COBIT 5 | 100% | 75% | NETAPP NS0 182 | 50% | 50% |
| ISQI CTFL | 100% | 75% | NETSUITE SUITEFOUNDATION CERTIFICATION EXAM | 50% | 50% |
| LPI 010 160 | 100% | 75% | NI CLAD | 50% | 50% |
| MICROSOFT 70 347 | 100% | 75% | NUTANIX NCP | 50% | 50% |
| MICROSOFT AZ 305 | 100% | 75% | PALO ALTO NETWORKS ACE | 50% | 50% |
| PALO ALTO NETWORKS PCCSA | 100% | 75% | PALO ALTO NETWORKS PCDRA | 50% | 50% |
| SIX SIGMA LSSYB | 100% | 75% | PEGASYSTEMS PEGAPCDC80V1 | 50% | 50% |
| SOA S9003 | 100% | 75% | RIVERBED 830 01 | 50% | 50% |
| TEST PREP GED SECTION 4 LANGUAGE ARTS READING | 100% | 75% | SALESFORCE CERTIFIED DATA ARCHITECT | 50% | 50% |
| WORLDATWORK C8 | 100% | 75% | SALESFORCE CERTIFIED OMNISTUDIO DEVELOPER | 50% | 50% |
| HRCI SPHR | 100% | 70% | SALESFORCE FIELD SERVICE LIGHTNING CONSULTANT | 50% | 50% |
| MICROSOFT 70 680 | 100% | 70% | SAP C S4CFI 2202 | 50% | 50% |
| ACFE CFE | 100% | 67% | SCRUM PSM II | 50% | 50% |
| MICROSOFT AZ 304 | 100% | 67% | SIX SIGMA ICYB | 50% | 50% |
| MICROSOFT SC 100 | 100% | 67% | SPLUNK SPLK 2002 | 50% | 50% |
| ALCATEL LUCENT 4A0 104 | 100% | 60% | SPLUNK SPLK 3001 | 50% | 50% |
| COMPTIA CAS 002 | 100% | 60% | SYMANTEC 250 428 | 50% | 50% |
| COMPTIA FC0 U51 | 100% | 60% | THE OPEN GROUP OG0 061 | 50% | 50% |
| F5 301B | 100% | 60% | THE OPEN GROUP OG0 092 | 50% | 50% |
| TEST PREP LSAT SECTION 2 READING COMPREHENSION | 100% | 60% | VERITAS VCS 323 | 50% | 50% |
| AVAYA 7004 | 100% | 50% | VERSA NETWORKS VNX100 | 50% | 50% |
| AVAYA 7591X | 100% | 50% | VMWARE 1V0 2120 | 50% | 50% |
| BCS ASTQB | 100% | 50% | VMWARE 2V0 2119 | 50% | 50% |
| CITRIX 1Y0 340 | 100% | 50% | VMWARE 2V0 2119D | 50% | 50% |
| CIW 1D0 520 | 100% | 50% | VMWARE 2V0 6120 | 50% | 50% |
| COMPTIA 220 1101 | 100% | 50% | VMWARE 2V0 6221 | 50% | 50% |
| COMPTIA CV1 003 | 100% | 50% | VMWARE 2V0 642 | 50% | 50% |
| CWNP PW0 071 | 100% | 50% | VMWARE 3V0 4220 | 50% | 50% |
| DELL DEA 5TT1 | 100% | 50% | VMWARE 5V0 23 20 | 50% | 50% |
| DELL DES 5121 | 100% | 50% | VMWARE 5V0 35 21 | 50% | 50% |
| ECCOUNCIL 312 49V8 | 100% | 50% | VMWARE 5V0 61 22 | 50% | 50% |
| ECCOUNCIL EC0 349 | 100% | 50% | VMWARE 5V0 6219 | 50% | 50% |
| EXIN EX0 002 | 100% | 50% | ZEND 200 710 | 50% | 50% |

| Certification | Turbo-GPT3.5 | GPT3 | Certification | Turbo-GPT3.5 | GPT3 |
|---|---|---|---|---|---|
| HITACHI HQT 4180 | 100% | 50% | GIAC GISF | 50% | 40% |
| ISTQB CTAL TM | 100% | 50% | HUAWEI H12 224 | 50% | 40% |
| MICROSOFT 70 537 | 100% | 50% | SANS SEC504 | 50% | 40% |
| MICROSOFT MB 600 | 100% | 50% | TEST PREP CFA LEVEL 3 | 50% | 40% |
| MICROSOFT MB2 712 | 100% | 50% | ISC CAP | 50% | 30% |
| MICROSOFT PL 100 | 100% | 50% | AIWMI CCRA | 50% | 25% |
| SAP C HANATEC 17 | 100% | 50% | ALCATEL LUCENT 4A0 M02 | 50% | 25% |
| SAP E ACTCLD 21 | 100% | 50% | APPIAN ACD200 | 50% | 25% |
| SAS INSTITUTE A00 250 | 100% | 50% | ARUBA ACMP 64 | 50% | 25% |
| SIX SIGMA LSSWB | 100% | 50% | AVAYA 3301 | 50% | 25% |
| SNIA S10 210 | 100% | 50% | AVAYA 3304 | 50% | 25% |
| SOA S9001 | 100% | 50% | AVAYA 6209 | 50% | 25% |
| TEST PREP CBEST SECTION 2 READING | 100% | 50% | AVAYA 71200X | 50% | 25% |
| VERITAS VCS 413 | 100% | 50% | AVAYA 71201X | 50% | 25% |
| VMWARE 2V0 5121 | 100% | 50% | AVAYA 7141X | 50% | 25% |
| ECCOUNCIL 312 49V10 | 100% | 40% | AVAYA 7304 | 50% | 25% |
| EXIN ITILF | 100% | 40% | BACB BCBA | 50% | 25% |
| LINUX FOUNDATION LFCS | 100% | 40% | CITRIX 1Y0 341 | 50% | 25% |
| MICROSOFT AZ 720 | 100% | 33% | CITRIX 1Y0 401 | 50% | 25% |
| MICROSOFT PL 400 | 100% | 33% | COMPTIA DA0 001 | 50% | 25% |
| VMWARE 1V0 642 | 100% | 33% | DATABRICKS CERTIFIED ASSOCIATE DEVELOPER FOR APACHE SPARK | 50% | 25% |
| ASQ CQA | 100% | 25% | DELL DCPPE 200 | 50% | 25% |
| CIW 1D0 610 | 100% | 25% | DELL DES 1221 | 50% | 25% |
| DELL E05 001 | 100% | 25% | DELL DES 1241 | 50% | 25% |
| HP HPE0 J74 | 100% | 25% | DELL DES 9131 | 50% | 25% |
| LPI 101 400 | 100% | 25% | DELL E20 020 | 50% | 25% |
| MICROSOFT 70 535 | 100% | 25% | DELL E20 575 | 50% | 25% |
| MICROSOFT 70 767 | 100% | 25% | DELL E20 655 | 50% | 25% |
| MICROSOFT 70 776 | 100% | 25% | DELL E20 920 | 50% | 25% |
| MICROSOFT 70 778 | 100% | 25% | ECCOUNCIL 312 76 | 50% | 25% |
| MICROSOFT MB 340 | 100% | 25% | FORTINET FORTISANDBOX | 50% | 25% |
| MICROSOFT MS 301 | 100% | 25% | FORTINET NSE5 FMG 6 4 | 50% | 25% |
| NETAPP NS0 171 | 100% | 25% | FORTINET NSE6 FAC 6 1 | 50% | 25% |
| SALESFORCE CERTIFIED SHARING AND VISIBILITY ARCHITECT | 100% | 25% | FORTINET NSE8 810 | 50% | 25% |
| SAS INSTITUTE A00 212 | 100% | 25% | FORTINET NSE8 811 | 50% | 25% |

| Certification | Turbo-GPT3.5 | GPT3 | Certification | Turbo-GPT3.5 | GPT3 |
|---|---|---|---|---|---|
| SAS INSTITUTE A00 240 | 100% | 25% | GENESYS CIC 101 01 | 50% | 25% |
| DELL E20 357 | 100% | 0% | GOOGLE PROFESSIONAL CLOUD DATABASE ENGINEER | 50% | 25% |
| MICROSOFT 70 331 | 100% | 0% | GUIDANCE SOFTWARE GD0 100 | 50% | 25% |
| MICROSOFT 70 466 | 100% | 0% | HP HPE0 J79 | 50% | 25% |
| MICROSOFT 70 705 | 100% | 0% | HP HPE0 S22 | 50% | 25% |
| MICROSOFT 98 367 | 100% | 0% | HP HPE6 A42 | 50% | 25% |
| MICROSOFT AZ 202 | 100% | 0% | ISACA CCAK | 50% | 25% |
| MICROSOFT MB 910 | 100% | 0% | ISTQB CTFL 2018 | 50% | 25% |
| MICROSOFT MB6 894 | 100% | 0% | LPI 202 450 | 50% | 25% |
| SOLARWINDS SCP 500 | 100% | 0% | MAGENTO M70 201 | 50% | 25% |
| VMWARE 2V0 2120 | 100% | 0% | META FACEBOOK 100 101 | 50% | 25% |
| VMWARE 2V0 631 | 100% | 0% | MICROSOFT 70 243 | 50% | 25% |
| ISC CCSP | 90% | 80% | MICROSOFT 70 346 | 50% | 25% |
| ALCATEL LUCENT 4A0 100 | 90% | 70% | MICROSOFT 70 765 | 50% | 25% |
| APICS CSCP | 90% | 70% | MICROSOFT 70 774 | 50% | 25% |
| ASIS ASIS CPP | 90% | 70% | MICROSOFT AZ 100 | 50% | 25% |
| COMPTIA SY0 401 | 90% | 70% | MICROSOFT MB 310 | 50% | 25% |
| MICROSOFT MS 100 | 90% | 70% | NETAPP NS0 155 | 50% | 25% |
| PEOPLECERT 58 | 90% | 70% | NETAPP NS0 162 | 50% | 25% |
| TEST PREP NCLEX RN | 90% | 70% | NETAPP NS0 175 | 50% | 25% |
| ACSM 010 111 | 90% | 60% | NETAPP NS0 191 | 50% | 25% |
| ECCOUNCIL 712 50 | 90% | 60% | NFPA CFPS | 50% | 25% |
| EXIN EX0 001 | 90% | 60% | NUTANIX NCA | 50% | 25% |
| TEST PREP MCAT TEST | 90% | 60% | PALO ALTO NETWORKS PCCSE | 50% | 25% |
| TEST PREP USMLE | 90% | 60% | PEGASYSTEMS PEGACPBA73V1 | 50% | 25% |
| ISACA CRISC | 90% | 50% | SALESFORCE CERTIFIED IDENTITY AND ACCESS MANAGEMENT DESIGNER | 50% | 25% |
| SIX SIGMA ICBB | 90% | 50% | SALESFORCE CERTIFIED INDUSTRIES CPQ DEVELOPER | 50% | 25% |
| SIX SIGMA LSSBB | 90% | 50% | SALESFORCE FIELD SERVICE CONSULTANT | 50% | 25% |
| ECCOUNCIL 312 49 | 90% | 20% | SERVICENOW CAS PA | 50% | 25% |
| COMPTIA N10 008 | 89% | 89% | SERVICENOW CIS APM | 50% | 25% |
| MICROSOFT MD 100 | 89% | 78% | SERVICENOW CIS DISCOVERY | 50% | 25% |
| COMPTIA N10 006 | 89% | 67% | SERVICENOW CIS SIR | 50% | 25% |
| MICROSOFT AZ 204 | 88% | 88% | TEST PREP ACLS | 50% | 25% |

| Certification | Turbo-GPT3.5 | GPT3 | Certification | Turbo-GPT3.5 | GPT3 |
|---|---|---|---|---|---|
| MICROSOFT MS 500 | 88% | 88% | TEST PREP GED SECTION 1 SOCIAL STUDIES | 50% | 25% |
| MICROSOFT MS 101 | 88% | 63% | TEST PREP HESI A2 | 50% | 25% |
| COMPTIA CS0 001 | 86% | 57% | TEST PREP RPFT | 50% | 25% |
| TEST PREP CGFM | 80% | 100% | VERITAS VCS 277 | 50% | 25% |
| ECCOUNCIL 312 50V10 | 80% | 90% | VMWARE 1V0 601 | 50% | 25% |
| SIX SIGMA LSSMBB | 80% | 90% | VMWARE 1V0 605 | 50% | 25% |
| AHIMA RHIA | 80% | 80% | VMWARE 2V0 2119 PSE | 50% | 25% |
| AMAZON AWS CERTIFIED MACHINE LEARNING SPECIALTY | 80% | 80% | VMWARE 2V0 3119 | 50% | 25% |
| AMAZON AWS CERTIFIED SYSOPS ADMINISTRATOR ASSOCIATE | 80% | 80% | VMWARE 2V0 4119 | 50% | 25% |
| CHECKPOINT 156 215 81 | 80% | 80% | VMWARE 2V0 72 22 | 50% | 25% |
| COMPTIA FC0 U61 | 80% | 80% | VMWARE 2V0 751 | 50% | 25% |
| MICROSOFT 70 697 | 80% | 80% | DELL E20 593 | 50% | 20% |
| MICROSOFT AZ 500 | 80% | 80% | SALESFORCE CERTIFIED PLATFORM DEVELOPER II | 50% | 20% |
| SCRUM PSM I | 80% | 80% | AMAZON AWS CERTIFIED BIG DATA SPECIALTY | 50% | 0% |
| CHECKPOINT 156 21577 | 80% | 70% | AVAYA 3002 | 50% | 0% |
| TEST PREP GMAT SECTION 3 | 80% | 70% | AVAYA 7130X | 50% | 0% |
| TEST PREP GMAT SECTION 3 VERBAL ABILITY | 80% | 70% | AVAYA 7492X | 50% | 0% |
| THE OPEN GROUP OG0 091 | 80% | 70% | AVAYA 76940X | 50% | 0% |
| COMPTIA 220 901 | 80% | 60% | CHECKPOINT 156 585 | 50% | 0% |
| COMPTIA SK0 005 | 80% | 60% | DELL DES 1D11 | 50% | 0% |
| GIAC GCIH | 80% | 60% | FORTINET NSE4 FGT 70 | 50% | 0% |
| GOOGLE PROFESSIONAL CLOUD ARCHITECT | 80% | 60% | FORTINET NSE6 FML 538 | 50% | 0% |
| HASHICORP TERRAFORM ASSOCIATE | 80% | 60% | FORTINET NSE6 FWB 61 | 50% | 0% |
| PMI PMI SP | 80% | 60% | FORTINET NSE6 FWF 64 | 50% | 0% |
| SALESFORCE ADM 211 | 80% | 60% | HITACHI HCE 3700 | 50% | 0% |
| TEST PREP CDL | 80% | 60% | HP HPE0 J68 | 50% | 0% |
| AMAZON AWS CERTIFIED SECURITY SPECIALTY | 80% | 50% | HP HPE0 J75 | 50% | 0% |
| COMPTIA CV0 001 | 80% | 50% | HP HPE0 Y53 | 50% | 0% |
| COMPTIA PK0 004 | 80% | 50% | MICROSOFT 70 341 | 50% | 0% |

| Certification | Turbo-GPT3.5 | GPT3 | Certification | Turbo-GPT3.5 | GPT3 |
|---|---|---|---|---|---|
| COMPTIA TK0 201 | 80% | 50% | MICROSOFT AZ 102 | 50% | 0% |
| GAQM APM 001 | 80% | 50% | MICROSOFT MB2 714 | 50% | 0% |
| GAQM CLSSBB | 80% | 50% | MICROSOFT MB2 877 | 50% | 0% |
| GOOGLE ADWORDS FUNDAMENTALS | 80% | 50% | NOKIA BL0 100 | 50% | 0% |
| GOOGLE INDIVIDUAL QUALIFICATION | 80% | 50% | PEGASYSTEMS PEGACPMC74V1 | 50% | 0% |
| SALESFORCE CRT 450 | 80% | 50% | PEGASYSTEMS PEGAPCSA80V1 2019 | 50% | 0% |
| TEST PREP ACT TEST | 80% | 50% | PEGASYSTEMS PEGAPCSSA87V1 | 50% | 0% |
| BLUE COAT BCCPA | 80% | 40% | SAP C SAC 2221 | 50% | 0% |
| ISC CISSP ISSMP | 80% | 40% | SAP C TS450 2020 | 50% | 0% |
| MICROSOFT DP 201 | 80% | 40% | SAP C TSCM62 67 | 50% | 0% |
| PMI CAPM | 80% | 40% | SERVICENOW CIS CSM | 50% | 0% |
| TEST PREP MCQS | 80% | 40% | SNOWFLAKE SNOWPRO ADVANCED ARCHITECT | 50% | 0% |
| TEST PREP SAT SECTION 1 CRITICAL READING | 80% | 40% | SOA S9002 | 50% | 0% |
| AMAZON AWS CERTIFIED CLOUD PRACTITIONER | 80% | 30% | VEEAM VMCE V9 | 50% | 0% |
| ALCATEL LUCENT 4A0 103 | 80% | 20% | VMWARE 1V0 701 | 50% | 0% |
| BLOCKCHAIN CBSA | 80% | 20% | VMWARE 2V0 731 | 50% | 0% |
| GIAC GSEC | 80% | 20% | VMWARE 2VB 601 | 50% | 0% |
| HUAWEI H12 221 | 80% | 20% | VMWARE 5V0 3219 | 50% | 0% |
| ISC CISSP ISSEP | 80% | 20% | MICROSOFT DP 200 | 43% | 43% |
| MICROSOFT 70 741 | 80% | 20% | SYMANTEC 250 513 | 40% | 70% |
| MICROSOFT 70 761 | 80% | 20% | ABA CTFA | 40% | 60% |
| MICROSOFT MD 101 | 78% | 67% | ALCATEL LUCENT 4A0 101 | 40% | 60% |
| ASQ CQE | 75% | 100% | BACB BCABA | 40% | 60% |
| ENGLISH TEST PREPARATION TOEFL SENTENCE CORRECTION | 75% | 100% | ECCOUNCIL 412 79V8 | 40% | 60% |
| FORTINET NSE5 FAZ 60 | 75% | 100% | MICROSOFT 70 744 | 40% | 60% |
| GOOGLE ASSOCIATE ANDROID DEVELOPER | 75% | 100% | SALESFORCE CERTIFIED ADVANCED ADMINISTRATOR | 40% | 60% |
| HP HPE2 E71 | 75% | 100% | SALESFORCE CERTIFIED SALES CLOUD CONSULTANT | 40% | 60% |
| IIBA ECBA | 75% | 100% | TEST PREP CFA LEVEL 2 | 40% | 60% |
| NETAPP NS0 145 | 75% | 100% | TEST PREP NET | 40% | 60% |
| PALO ALTO NETWORKS PSE SASE | 75% | 100% | AMAZON AWS SYSOPS | 40% | 40% |

| Certification | Turbo-GPT3.5 | GPT3 | Certification | Turbo-GPT3.5 | GPT3 |
|---|---|---|---|---|---|
| SALESFORCE CERTIFIED MARKETING CLOUD ADMINISTRATOR | 75% | 100% | CHECKPOINT 156 31580 | 40% | 40% |
| VMWARE 2V0 7121 | 75% | 100% | DELL E20 260 | 40% | 40% |
| A10 NETWORKS A10 CERTIFIED PROFESSIONAL SYSTEM ADMINISTRATION 4 | 75% | 75% | HP HPE6 A29 | 40% | 40% |
| AAFM INDIA CWM LEVEL 1 | 75% | 75% | PMI PFMP | 40% | 40% |
| AVAYA 7003 | 75% | 75% | SAS INSTITUTE A00 211 | 40% | 40% |
| AVAYA 7392X | 75% | 75% | TEST PREP PTCE | 40% | 40% |
| BLOCKCHAIN CBBF | 75% | 75% | TEST PREP SAT TEST | 40% | 40% |
| CHECKPOINT 156 315 81 | 75% | 75% | TEST PREP TCLEOSE | 40% | 40% |
| CITRIX 1Y0 231 | 75% | 75% | VMWARE 2V0 621 | 40% | 40% |
| CIW 1D0 541 | 75% | 75% | VMWARE 2V0 621D | 40% | 40% |
| COMPTIA CLO 002 | 75% | 75% | APICS CLTD | 40% | 20% |
| COMPTIA XK0 004 | 75% | 75% | BLUE COAT BCCPP | 40% | 20% |
| CROWDSTRIKE CCFA | 75% | 75% | F5 101 | 40% | 20% |
| CWNP CWAP 402 | 75% | 75% | FINRA SERIES 7 | 40% | 20% |
| CWNP CWNA 107 | 75% | 75% | GIAC GSSP JAVA | 40% | 20% |
| CWNP CWNA 108 | 75% | 75% | HRCI GPHR | 40% | 20% |
| DELL DEA 2TT3 | 75% | 75% | IAPP CIPM | 40% | 20% |
| DELL E10 002 | 75% | 75% | ISACA CDPSE | 40% | 20% |
| ECCOUNCIL 312 85 | 75% | 75% | SALESFORCE CERTIFIED MARKETING CLOUD EMAIL SPECIALIST | 40% | 20% |
| ENGLISH TEST PREPARATION TOEFL SENTENCE COMPLETION | 75% | 75% | TEST PREP PCAT | 40% | 20% |
| EXIN ISFS | 75% | 75% | CHECKPOINT 156 91577 | 40% | 10% |
| GENESYS GCP GC IMP | 75% | 75% | DELL E20 555 | 40% | 0% |
| GENESYS GCP GCX | 75% | 75% | RIVERBED 101 01 | 40% | 0% |
| GOOGLE PROFESSIONAL CLOUD DEVOPS ENGINEER | 75% | 75% | MICROSOFT MB 210 | 38% | 50% |
| GOOGLE PROFESSIONAL CLOUD NETWORK ENGINEER | 75% | 75% | DELL DES 1121 | 33% | 67% |
| GOOGLE SHOPPING ADVERTISING | 75% | 75% | HP HP0 Y47 | 33% | 67% |
| HP HPE0 S46 | 75% | 75% | HP HPE2 T37 | 33% | 67% |

| Certification | Turbo-GPT3.5 | GPT3 | Certification | Turbo-GPT3.5 | GPT3 |
|---|---|---|---|---|---|
| HP HPE6 A70 | 75% | 75% | MICROSOFT 70 742 | 33% | 67% |
| HP HPE6 A73 | 75% | 75% | MICROSOFT MS 740 | 33% | 67% |
| MAGENTO M70 301 | 75% | 75% | MICROSOFT SC 200 | 33% | 67% |
| MICROSOFT 70 463 | 75% | 75% | CHECKPOINT 156 31577 | 33% | 44% |
| MICROSOFT 70 685 | 75% | 75% | COMPTIA LX0 103 | 33% | 33% |
| MICROSOFT 70 773 | 75% | 75% | CYBERARK PAM CDE RECERT | 33% | 33% |
| MICROSOFT 74 409 | 75% | 75% | DELL DES 6332 | 33% | 33% |
| MICROSOFT DP 300 | 75% | 75% | DELL DES DD33 | 33% | 33% |
| MICROSOFT MB 230 | 75% | 75% | HP HP2 B149 | 33% | 33% |
| MICROSOFT MB2 708 | 75% | 75% | HP HPE6 A72 | 33% | 33% |
| MICROSOFT SC 300 | 75% | 75% | MICROSOFT 70 462 | 33% | 33% |
| NADCA ASCS | 75% | 75% | MICROSOFT 70 480 | 33% | 33% |
| NETAPP NS0 180 | 75% | 75% | MICROSOFT 70 496 | 33% | 33% |
| NETAPP NS0 194 | 75% | 75% | MICROSOFT DP 203 | 33% | 33% |
| NOVELL 050 733 | 75% | 75% | MICROSOFT MB 220 | 33% | 33% |
| PALO ALTO NETWORKS PSE STRATA | 75% | 75% | MICROSOFT MS 600 | 33% | 33% |
| SALESFORCE CERTIFIED BUSINESS ANALYST | 75% | 75% | MICROSOFT PL 900 | 33% | 33% |
| SAP C TAW12 750 | 75% | 75% | PEGASYSTEMS PEGACSA72V1 | 33% | 33% |
| SCALED AGILE SA | 75% | 75% | SALESFORCE DEV 401 | 33% | 33% |
| SIX SIGMA ICGB | 75% | 75% | VMWARE 2V0 602 | 33% | 33% |
| SIX SIGMA LSSGB | 75% | 75% | VMWARE 3V0 624 | 33% | 33% |
| SYMANTEC 250 430 | 75% | 75% | DELL E20 307 | 33% | 0% |
| TEST PREP TEAS SECTION 2 SENTENCE CORRECTION | 75% | 75% | FORTINET NSE5 FAZ 62 | 33% | 0% |
| VMWARE 1V0 603 | 75% | 75% | HP HPE0 S58 | 33% | 0% |
| VMWARE 2V0 6119 | 75% | 75% | MICROSOFT 70 473 | 33% | 0% |
| VMWARE 2V0 620 | 75% | 75% | MICROSOFT 70 695 | 33% | 0% |
| WORLDATWORK T1 GR1 | 75% | 75% | MICROSOFT 70 713 | 33% | 0% |
| AMAZON AWS CERTIFIED ADVANCED NETWORKING SPECIALTY ANS C01 | 75% | 50% | MICROSOFT 98 369 | 33% | 0% |
| ANDROIDATC AND 403 | 75% | 50% | PEGASYSTEMS PEGACRSA80V1 | 33% | 0% |
| APPLE MAC 16A | 75% | 50% | HUAWEI H13 629 | 30% | 40% |
| AVAYA 3107 | 75% | 50% | ISACA CGEIT | 30% | 30% |
| AVAYA 7120X | 75% | 50% | GIAC GPEN | 30% | 10% |
| AVAYA 7593X | 75% | 50% | AVAYA 6211 | 25% | 75% |

| Certification | Turbo-GPT3.5 | GPT3 | Certification | Turbo-GPT3.5 | GPT3 |
|---|---|---|---|---|---|
| BCS TM12 | 75% | 50% | AVAYA 7230X | 25% | 75% |
| BLOCKCHAIN CBDE | 75% | 50% | DELL E20 893 | 25% | 75% |
| CHECKPOINT 156 110 | 75% | 50% | GENESYS GCP GC REP | 25% | 75% |
| CIW 1D0 571 | 75% | 50% | HP HP2 B148 | 25% | 75% |
| COMPTIA LX0 104 | 75% | 50% | HP HPE0 S52 | 25% | 75% |
| ECCOUNCIL 212 89 | 75% | 50% | HP HPE6 A49 | 25% | 75% |
| EXIN EX0 105 | 75% | 50% | HUAWEI H19 301 | 25% | 75% |
| EXIN PR2P | 75% | 50% | CHECKPOINT 156 91580 | 25% | 50% |
| FORTINET NSE4 FGT 62 | 75% | 50% | CITRIX 1Y0 204 | 25% | 50% |
| FORTINET NSE5 FMG 7 0 | 75% | 50% | CITRIX 1Y0 311 | 25% | 50% |
| FORTINET NSE6 | 75% | 50% | CIW 1D0 621 | 25% | 50% |
| FORTINET NSE6 FWB 560 | 75% | 50% | DELL DES 1721 | 25% | 50% |
| FORTINET NSE7 EFW 7 0 | 75% | 50% | DELL DES 1B31 | 25% | 50% |
| FORTINET NSE7 PBC 64 | 75% | 50% | DELL DNDNS 200 | 25% | 50% |
| GIAC GCPM | 75% | 50% | ENGLISH TEST PREPARATION TOEFL READING COMPREHENSION | 25% | 50% |
| GOOGLE MOBILE ADVERTISING | 75% | 50% | EXIN ISMP | 25% | 50% |
| GOOGLE PROFESSIONAL COLLABORATION ENGINEER | 75% | 50% | FORTINET NSE7 | 25% | 50% |
| GOOGLE VIDEO ADVERTISING | 75% | 50% | GENESYS GE0 806 | 25% | 50% |
| HP HPE0 S37 | 75% | 50% | HP HPE2 E72 | 25% | 50% |
| HP HPE2 E69 | 75% | 50% | HP HPE6 A07 | 25% | 50% |
| HRCI PHR | 75% | 50% | INFOR IOS 252 | 25% | 50% |
| INFOR M3 123 | 75% | 50% | ISQI CTAL TA | 25% | 50% |
| ISTQB ATM | 75% | 50% | MCAFEE MA0 101 | 25% | 50% |
| ISTQB ATTA | 75% | 50% | MICROSOFT 74 343 | 25% | 50% |
| LPI 010 150 | 75% | 50% | MICROSOFT MB 240 | 25% | 50% |
| MICROSOFT 70 483 | 75% | 50% | NETAPP NS0 502 | 25% | 50% |
| MICROSOFT 70 486 | 75% | 50% | SALESFORCE CERTIFIED HEROKU ARCHITECTURE DESIGNER | 25% | 50% |
| MICROSOFT 70 735 | 75% | 50% | SALESFORCE CERTIFIED SERVICE CLOUD CONSULTANT | 25% | 50% |
| MICROSOFT 70 768 | 75% | 50% | SERVICENOW CIS EM | 25% | 50% |
| MICROSOFT 77 602 | 75% | 50% | SERVICENOW CIS ITSM | 25% | 50% |
| MICROSOFT 77 881 | 75% | 50% | SOA S9009 | 25% | 50% |

| Certification | Turbo-GPT3.5 | GPT3 | Certification | Turbo-GPT3.5 | GPT3 |
|---|---|---|---|---|---|
| MICROSOFT 98 368 | 75% | 50% | VMWARE 5V0 2120 | 25% | 50% |
| MICROSOFT AZ 301 | 75% | 50% | VMWARE 5V0 3221 | 25% | 50% |
| MICROSOFT DA 100 | 75% | 50% | AHLEI AHLEI CHA | 25% | 25% |
| NCINO 201 COMMERCIAL BANKING FUNCTIONAL | 75% | 50% | ALCATEL LUCENT 4A0 102 | 25% | 25% |
| NETAPP NS0 507 | 75% | 50% | AMAZON AWS CERTIFIED ALEXA SKILL BUILDER SPECIALTY | 25% | 25% |
| NOKIA 4A0 116 | 75% | 50% | AMAZON AWS CERTIFIED DATA ANALYTICS SPECIALTY | 25% | 25% |
| NOKIA 4A0 AI1 | 75% | 50% | AVAYA 3108 | 25% | 25% |
| QLIKVIEW QV DEVELOPER 01 | 75% | 50% | AVAYA 3309 | 25% | 25% |
| REDHAT EX407 | 75% | 50% | AVAYA 7491X | 25% | 25% |
| SALESFORCE CERTIFIED COMMUNITY CLOUD CONSULTANT | 75% | 50% | AVAYA 7495X | 25% | 25% |
| SALESFORCE CERTIFIED DEVELOPMENT LIFECYCLE AND DEPLOYMENT DESIGNER | 75% | 50% | AVAYA 75940X | 25% | 25% |
| SALESFORCE CRT 160 | 75% | 50% | AVAYA 7765X | 25% | 25% |
| SAP C SECAUTH 20 | 75% | 50% | BLUE PRISM APD01 | 25% | 25% |
| SERVICENOW CAD | 75% | 50% | CERTNEXUS CFR 310 | 25% | 25% |
| SERVICENOW CIS PPM | 75% | 50% | CITRIX 1Y0 240 | 25% | 25% |
| SERVICENOW CIS VRM | 75% | 50% | CITRIX 1Y0 402 | 25% | 25% |
| SOA S9008 | 75% | 50% | CYBERARK CAU301 | 25% | 25% |
| TABLEAU TDS C01 | 75% | 50% | CYBERARK PAM SEN | 25% | 25% |
| TEST PREP GED SECTION 3 SCIENCE | 75% | 50% | DELL DSDSC 200 | 25% | 25% |
| TEST PREP NAPLEX | 75% | 50% | DELL E20 562 | 25% | 25% |
| TEST PREP NCE | 75% | 50% | DELL E20 585 | 25% | 25% |
| TEST PREP TEAS SECTION 1 READING COMPREHENSION | 75% | 50% | DELL E20 891 | 25% | 25% |
| THE OPEN GROUP OG0 023 | 75% | 50% | F5 201 | 25% | 25% |
| VEEAM VMCE 2021 | 75% | 50% | FORTINET NSE5 FAZ 7 0 | 25% | 25% |
| VEEAM VMCE2020 | 75% | 50% | FORTINET NSE5 FMG 60 | 25% | 25% |
| VMWARE 2V0 33 22 | 75% | 50% | FORTINET NSE8 | 25% | 25% |
| VMWARE 2V0 4120 | 75% | 50% | GENESYS GE0 807 | 25% | 25% |
| APPLE 9L0 422 | 75% | 25% | GIAC GASF | 25% | 25% |

| Certification | Turbo-GPT3.5 | GPT3 | Certification | Turbo-GPT3.5 | GPT3 |
|---|---|---|---|---|---|
| ASQ CQIA | 75% | 25% | GOOGLE ASSOCIATE CLOUD ENGINEER | 25% | 25% |
| AVAYA 3314 | 75% | 25% | HP HP0 Y50 | 25% | 25% |
| AVAYA 78200X | 75% | 25% | HP HPE0 J80 | 25% | 25% |
| BLUE PRISM ATA02 | 75% | 25% | HP HPE6 A41 | 25% | 25% |
| CA TECHNOLOGIES CAT 340 | 75% | 25% | HP HPE6 A44 | 25% | 25% |
| CITRIX 1Y0 203 | 75% | 25% | HUAWEI H12 311 | 25% | 25% |
| CITRIX 1Y0 230 | 75% | 25% | ISTQB ATA | 25% | 25% |
| CWNP CWSP 205 | 75% | 25% | ISTQB CT TAE | 25% | 25% |
| DELL DES 1423 | 75% | 25% | MCAFEE MA0 104 | 25% | 25% |
| DELL DES 3611 | 75% | 25% | MICROSOFT 70 703 | 25% | 25% |
| DELL DES 4421 | 75% | 25% | MICROSOFT 77 884 | 25% | 25% |
| DELL E20 542 | 75% | 25% | MICROSOFT AZ 303 | 25% | 25% |
| DELL E20 594 | 75% | 25% | MICROSOFT MB2 711 | 25% | 25% |
| ECCOUNCIL ECSAV10 | 75% | 25% | MICROSOFT MB2 719 | 25% | 25% |
| ENGLISH TEST PREPARATION COMPLETE IELTS GUIDE | 75% | 25% | MICROSOFT MB6 895 | 25% | 25% |
| FORTINET NSE4 FGT 64 | 75% | 25% | MICROSOFT MS 202 | 25% | 25% |
| FORTINET NSE5 FMG 54 | 75% | 25% | NETAPP NS0 183 | 25% | 25% |
| GOOGLE PROFESSIONAL CLOUD SECURITY ENGINEER | 75% | 25% | NETAPP NS0 520 | 25% | 25% |
| HP HPE0 J77 | 75% | 25% | NUTANIX NCM MCI | 25% | 25% |
| HP HPE0 S47 | 75% | 25% | PEGASYSTEMS PEGACSSA74V1 | 25% | 25% |
| HP HPE6 A45 | 75% | 25% | PEGASYSTEMS PEGAPCSA86V1 | 25% | 25% |
| HP HPE6 A48 | 75% | 25% | PYTHON INSTITUTE PCAP | 25% | 25% |
| IAPP CIPP US | 75% | 25% | RIVERBED 499 01 | 25% | 25% |
| LPI 304 200 | 75% | 25% | SALESFORCE CERTIFIED EINSTEIN ANALYTICS AND DISCOVERY CONSULTANT | 25% | 25% |
| MICROSOFT 70 497 | 75% | 25% | SALESFORCE CERTIFIED MARKETING CLOUD CONSULTANT | 25% | 25% |
| MICROSOFT 70 686 | 75% | 25% | SALESFORCE CERTIFIED TABLEAU CRM AND EINSTEIN DISCOVERY CONSULTANT | 25% | 25% |
| MICROSOFT 70 745 | 75% | 25% | SAP C TFIN52 67 | 25% | 25% |
| MICROSOFT 70 775 | 75% | 25% | SAP C TS452 2021 | 25% | 25% |
| MICROSOFT 70 779 | 75% | 25% | SAP C TS462 2020 | 25% | 25% |

| Certification | Turbo-GPT3.5 | GPT3 | Certification | Turbo-GPT3.5 | GPT3 |
|---|---|---|---|---|---|
| MICROSOFT AZ 220 | 75% | 25% | SAP P S4FIN 2021 | 25% | 25% |
| MICROSOFT AZ 700 | 75% | 25% | SERVICENOW CIS CPG | 25% | 25% |
| MICROSOFT MB2 707 | 75% | 25% | SERVICENOW CIS HAM | 25% | 25% |
| MICROSOFT MB6 896 | 75% | 25% | SITECORE NET DEVELOPER 10 | 25% | 25% |
| MICROSOFT MB6 898 | 75% | 25% | SYMANTEC 250 438 | 25% | 25% |
| MICROSOFT MS 300 | 75% | 25% | TEST PREP GED SECTION 2 LANGUAGE ARTS WRITING | 25% | 25% |
| NETAPP NS0 002 | 75% | 25% | VERITAS VCS 257 | 25% | 25% |
| NETAPP NS0 184 | 75% | 25% | VERITAS VCS 274 | 25% | 25% |
| PALO ALTO NETWORKS PCSAE | 75% | 25% | VERITAS VCS 275 | 25% | 25% |
| SALESFORCE DEV 450 | 75% | 25% | VERITAS VCS 322 | 25% | 25% |
| SPLUNK SPLK 1002 | 75% | 25% | VMEDU SCRUM MASTER CERTIFIED | 25% | 25% |
| TEST PREP AACD | 75% | 25% | VMWARE 1V0 31 21 | 25% | 25% |
| TEST PREP CBEST SECTION 1 MATH | 75% | 25% | VMWARE 2V0 0119 | 25% | 25% |
| TIBCO TB0 123 | 75% | 25% | VMWARE 2V0 3121 | 25% | 25% |
| UIPATH UIRPA | 75% | 25% | VMWARE 2V0 81 20 | 25% | 25% |
| VERITAS VCS 272 | 75% | 25% | VMWARE 5V0 2119 | 25% | 25% |
| VERITAS VCS 276 | 75% | 25% | WATCHGUARD ESSENTIALS | 25% | 25% |
| VERITAS VCS 279 | 75% | 25% | ATLASSIAN ACP 600 | 25% | 0% |
| VMWARE 1V0 71 21 | 75% | 25% | AVAYA 6210 | 25% | 0% |
| VMWARE 2V0 5119 | 75% | 25% | BLUE PRISM ASD01 | 25% | 0% |
| VMWARE 2V0 651 | 75% | 25% | CITRIX 1Y0 241 | 25% | 0% |
| VMWARE 5V0 1121 | 75% | 25% | CITRIX 1Y0 371 | 25% | 0% |
| VMWARE 5V0 2221 | 75% | 25% | CITRIX 1Y0 440 | 25% | 0% |
| ARISTA ACE A12 | 75% | 0% | CITRIX 1Y0 A20 | 25% | 0% |
| AVAYA 7220X | 75% | 0% | CWNP CWDP 302 | 25% | 0% |
| AVAYA 7893X | 75% | 0% | DELL DEE 1421 | 25% | 0% |
| AXIS COMMUNICATIONS AX0 100 | 75% | 0% | DELL DES 1D12 | 25% | 0% |
| DELL E20 007 | 75% | 0% | DELL DES 6321 | 25% | 0% |
| DELL E20 385 | 75% | 0% | FILEMAKER FM0 308 | 25% | 0% |
| FORTINET NSE5 FAZ 6 4 | 75% | 0% | FORTINET NSE5 FMG 62 | 25% | 0% |
| FORTINET NSE7 EFW 62 | 75% | 0% | FORTINET NSE6 FML 6 2 | 25% | 0% |
| FORTINET NSE7 EFW 64 | 75% | 0% | FORTINET NSE6 FML 6 4 | 25% | 0% |
| HP HPE0 S54 | 75% | 0% | FORTINET NSE7 OTS 64 | 25% | 0% |
| PEGASYSTEMS PEGAPCSSA80V1 2019 | 75% | 0% | GIAC GCED | 25% | 0% |
| SAP C CPI 14 | 75% | 0% | HP HPE0 J58 | 25% | 0% |

| Certification | Turbo-GPT3.5 | GPT3 | Certification | Turbo-GPT3.5 | GPT3 |
|---|---|---|---|---|---|
| SAP C TS413 2021 | 75% | 0% | HP HPE0 S57 | 25% | 0% |
| SPLUNK SPLK 2001 | 75% | 0% | MAGENTO M70 101 | 25% | 0% |
| TEST PREP PRAXIS MATHEMATICS SECTION | 75% | 0% | MAGENTO MAGENTO CERTIFIED PROFESSIONAL CLOUD DEVELOPER | 25% | 0% |
| VERITAS VCS 411 | 75% | 0% | MICROSOFT MB2 716 | 25% | 0% |
| COMPTIA SY0 501 | 71% | 43% | MICROSOFT MB2 718 | 25% | 0% |
| MICROSOFT 70 410 | 71% | 29% | MICROSOFT MB6 897 | 25% | 0% |
| AMAZON AWS CERTIFIED SOLUTIONS ARCHITECT ASSOCIATE SAA C02 | 70% | 80% | NETAPP NS0 160 | 25% | 0% |
| COMPTIA SK0 004 | 70% | 80% | NETAPP NS0 161 | 25% | 0% |
| COMPTIA N10 007 | 70% | 70% | NETAPP NS0 509 | 25% | 0% |
| FINRA SERIES 6 | 70% | 70% | PEGASYSTEMS PEGAPCRSA80V1 2019 | 25% | 0% |
| ISC CISSP | 70% | 70% | QLIKVIEW QSBA2018 | 25% | 0% |
| PMI PMP | 70% | 70% | QLIKVIEW QSDA2018 | 25% | 0% |
| ASQ CMQ OE | 70% | 60% | SALESFORCE CERTIFIED DEVELOPMENT LIFECYCLE AND DEPLOYMENT ARCHITECT | 25% | 0% |
| GIAC GISP | 70% | 60% | SALESFORCE CERTIFIED EXPERIENCE CLOUD CONSULTANT | 25% | 0% |
| ISACA CISM | 70% | 60% | SALESFORCE CERTIFIED INTEGRATION ARCHITECT | 25% | 0% |
| SERVICENOW CSA | 70% | 60% | SALESFORCE CERTIFIED SHARING AND VISIBILITY DESIGNER | 25% | 0% |
| AMAZON AWS CERTIFIED DEVELOPER ASSOCIATE | 70% | 50% | SAP C TPLM30 67 | 25% | 0% |
| GIAC GCIA | 70% | 50% | SAP C TS413 1909 | 25% | 0% |
| IIBA CBAP | 70% | 50% | SAP C TS4FI 2020 | 25% | 0% |
| SNOWFLAKE SNOWPRO CORE | 70% | 40% | SAP E HANAAW 17 | 25% | 0% |
| GIAC GSLC | 70% | 30% | SAS INSTITUTE A00 281 | 25% | 0% |
| ISC SSCP | 70% | 30% | SERVICENOW CIS RC | 25% | 0% |
| TEST PREP GED TEST | 70% | 30% | TEST PREP GED SECTION 5 MATHEMATICS | 25% | 0% |
| FORTINET NSE4 54 | 70% | 20% | TEST PREP WORKKEYS | 25% | 0% |
| RIVERBED 199 01 | 70% | 20% | VMWARE 2V0 3120 | 25% | 0% |
| BICSI RCDD 001 | 70% | 10% | VMWARE 3V0 3221 | 25% | 0% |
| CLOUDERA CCD 410 | 67% | 100% | VMWARE 3V0 732 | 25% | 0% |
| MICROSOFT AZ 300 | 67% | 100% | VMWARE 5V0 31 20 | 25% | 0% |

| Certification | Turbo-GPT3.5 | GPT3 | Certification | Turbo-GPT3.5 | GPT3 |
|---|---|---|---|---|---|
| MICROSOFT MB 500 | 67% | 100% | VMWARE 5V0 3419 | 25% | 0% |
| MICROSOFT AZ 400 | 67% | 89% | VMWARE 5V0 9120 | 25% | 0% |
| APPLE 9L0 066 | 67% | 67% | COMPTIA CS0 002 | 22% | 22% |
| DELL DEA 3TT2 | 67% | 67% | COMPTIA CLO 001 | 20% | 60% |
| ECCOUNCIL 312 50 | 67% | 67% | MICROSOFT 70 411 | 20% | 40% |
| ECCOUNCIL 312 50V8 | 67% | 67% | TEST PREP GRE SECTION 1 VERBAL | 20% | 40% |
| HP HPE0 S50 | 67% | 67% | DELL E20 329 | 20% | 30% |
| HP HPE0 V14 | 67% | 67% | ALCATEL LUCENT 4A0 107 | 20% | 20% |
| ISC CSSLP | 67% | 67% | AMAZON AWS CERTIFIED DATABASE SPECIALTY | 20% | 20% |
| LPI 300 100 | 67% | 67% | DELL E20 375 | 20% | 20% |
| LPI 303 200 | 67% | 67% | DELL E20 895 | 20% | 20% |
| MICROSOFT 70 461 | 67% | 67% | MICROSOFT 70 743 | 20% | 20% |
| MICROSOFT 77 888 | 67% | 67% | TEST PREP GMAT SECTION 2 QUANTITATIVE | 20% | 20% |
| MICROSOFT MB 800 | 67% | 67% | VMWARE 2V0 622 | 20% | 20% |
| MICROSOFT PL 300 | 67% | 67% | HUAWEI H11 851 | 20% | 0% |
| NOVELL 050 720 | 67% | 67% | SALESFORCE CERTIFIED PLATFORM APP BUILDER | 20% | 0% |
| ECCOUNCIL EC0 350 | 67% | 56% | TEST PREP NREMT | 20% | 0% |
| MICROSOFT 70 417 | 67% | 50% | COMPTIA CAS 003 | 14% | 29% |
| MICROSOFT MB 300 | 67% | 44% | TEST PREP GMAT SECTION 2 | 14% | 0% |
| APPIAN ACD100 | 67% | 33% | TEST PREP GMAT TEST | 11% | 33% |
| COMPTIA 220 1002 | 67% | 33% | SAP C | 11% | 22% |
| ECCOUNCIL 312 38 | 67% | 33% | TEST PREP LSAT SECTION 1 LOGICAL REASONING | 10% | 50% |
| LPI 102 400 | 67% | 33% | FORTINET NSE5 | 10% | 20% |
| MICROSOFT 70 764 | 67% | 33% | HUAWEI H12 211 | 10% | 20% |
| MICROSOFT 74 697 | 67% | 33% | MICROSOFT MS 200 | 0% | 75% |
| MICROSOFT MB 260 | 67% | 33% | MICROSOFT MS 220 | 0% | 67% |
| MICROSOFT MB 700 | 67% | 33% | AVAYA 7303 | 0% | 50% |
| MICROSOFT PL 500 | 67% | 33% | CITRIX 1Y0 312 | 0% | 50% |
| SERVICENOW CIS SAM | 67% | 33% | DELL E20 526 | 0% | 50% |
| VMWARE 5V0 2121 | 67% | 33% | FORTINET NSE5 FSM 5 2 | 0% | 50% |
| HP HPE6 A82 | 67% | 0% | HP HPE6 A15 | 0% | 50% |
| LPI 201 450 | 67% | 0% | MCAFEE MA0 100 | 0% | 50% |
| MICROSOFT 70 532 | 67% | 0% | MICROSOFT AZ 120 | 0% | 50% |
| COMPTIA CV0 002 | 63% | 50% | MICROSOFT AZ 302 | 0% | 50% |
| MICROSOFT 70 413 | 63% | 50% | MICROSOFT SC 400 | 0% | 50% |
| PALO ALTO NETWORKS PCNSA | 63% | 50% | NETAPP NS0 170 | 0% | 50% |
| TEST PREP CFA LEVEL 1 | 63% | 50% | SERVICENOW CIS SM | 0% | 50% |

| Certification | Turbo-GPT3.5 | GPT3 | Certification | Turbo-GPT3.5 | GPT3 |
|---|---|---|---|---|---|
| GIAC GPPA | 60% | 80% | SPLUNK SPLK 1003 | 0% | 50% |
| PALO ALTO NETWORKS PCNSE | 60% | 80% | AVAYA 7497X | 0% | 33% |
| BCS PRF | 60% | 60% | AVAYA 77200X | 0% | 25% |
| COMPTIA CAS 004 | 60% | 60% | CHECKPOINT 156 835 | 0% | 25% |
| CYBERARK CAU302 | 60% | 60% | DELL DEP 3CR1 | 0% | 25% |
| DELL E20 591 | 60% | 60% | DELL E20 559 | 0% | 25% |
| GIAC GCFA | 60% | 60% | FORTINET NSE4 FGT 7 2 | 0% | 25% |
| GOOGLE SEARCH ADVERTISING | 60% | 60% | FORTINET NSE6 FNC 85 | 0% | 25% |
| MICROSOFT 70 740 | 60% | 60% | HP HPE2 K42 | 0% | 25% |
| PALO ALTO NETWORKS PCCET | 60% | 60% | HP HPE6 A43 | 0% | 25% |
| PMI PGMP | 60% | 60% | HUAWEI H12 711 | 0% | 25% |
| SYMANTEC 250 315 | 60% | 60% | HUAWEI H13 611 | 0% | 25% |
| ZEND 200 550 | 60% | 60% | HUAWEI H13 612 | 0% | 25% |
| ABA CRCM | 60% | 50% | NETAPP NS0 158 | 0% | 25% |
| AHIP AHM 250 | 60% | 50% | NETAPP NS0 505 | 0% | 25% |
| COMPTIA SY0 601 | 60% | 50% | NUTANIX NCP MCA | 0% | 25% |
| PMI PMI RMP | 60% | 50% | PEGASYSTEMS PEGACSSA72V1 | 0% | 25% |
| SALESFORCE ADM 201 | 60% | 50% | SAP C THR81 2205 | 0% | 25% |
| AMAZON AWS DEVOPS ENGINEER PROFESSIONAL | 60% | 40% | SPLUNK SPLK 3002 | 0% | 25% |
| ASQ CSSBB | 60% | 40% | ALCATEL LUCENT 4A0 108 | 0% | 0% |
| CHECKPOINT 156 21580 | 60% | 40% | AMAZON AWS CERTIFIED SAP ON AWS SPECIALTY PAS C01 | 0% | 0% |
| COMPTIA CV0 003 | 60% | 40% | AVAYA 7241X | 0% | 0% |
| DELL E20 393 | 60% | 40% | AVAYA 7693X | 0% | 0% |
| GOOGLE DISPLAY ADVERTISING | 60% | 40% | AVAYA 7750X | 0% | 0% |
| MICROSOFT AZ 103 | 60% | 40% | COMPTIA PT0 001 | 0% | 0% |
| PMI PMI ACP | 60% | 40% | CYBERARK PAM DEF | 0% | 0% |
| RIVERBED 201 01 | 60% | 40% | DELL DES 1111 | 0% | 0% |
| RIVERBED 299 01 | 60% | 40% | DELL DES 1B21 | 0% | 0% |
| SALESFORCE CERTIFIED CPQ SPECIALIST | 60% | 40% | DELL DES 6322 | 0% | 0% |
| SAP C THR12 67 | 60% | 40% | DELL DSDPS 200 | 0% | 0% |
| SPLUNK SPLK 1001 | 60% | 40% | DELL E20 507 | 0% | 0% |
| TEST PREP ASVAB TEST | 60% | 40% | DELL E22 285 | 0% | 0% |
| TEST PREP CPA AUDITING AND ATTESTATION | 60% | 40% | EXIN DEVOPSF | 0% | 0% |

| Certification | Turbo-GPT3.5 | GPT3 | Certification | Turbo-GPT3.5 | GPT3 |
|---|---|---|---|---|---|
| TEST PREP LEED | 60% | 30% | HP HP2 Z34 | 0% | 0% |
| TEST PREP NCLEX PN | 60% | 30% | HP HPE6 A68 | 0% | 0% |
| AMAZON ANS C00 | 60% | 20% | HP HPE6 A79 | 0% | 0% |
| DELL E20 598 | 60% | 20% | HUAWEI H13 341 | 0% | 0% |
| TEST PREP NCMA | 60% | 20% | IIBA CCBA | 0% | 0% |
| VMWARE 2V0 622D | 60% | 20% | ISAQB CPSA F | 0% | 0% |
| FINRA SERIES 63 | 60% | 0% | LPI 701 100 | 0% | 0% |
| COMPTIA 220 1001 | 57% | 29% | MICROSOFT 70 333 | 0% | 0% |
| MICROSOFT 70 412 | 57% | 14% | MICROSOFT 70 345 | 0% | 0% |
| MICROSOFT AZ 900 | 56% | 89% | MICROSOFT 70 467 | 0% | 0% |
| TEST PREP HSPT TEST | 56% | 33% | MICROSOFT 70 498 | 0% | 0% |
| MICROSOFT 70 464 | 50% | 100% | MICROSOFT 70 533 | 0% | 0% |
| MICROSOFT MB 200 | 50% | 100% | MICROSOFT 98 383 | 0% | 0% |
| NCMA CPCM | 50% | 100% | MICROSOFT 98 388 | 0% | 0% |
| NOKIA 4A0 114 | 50% | 100% | MICROSOFT AZ 203 | 0% | 0% |
| NOKIA 4A0 205 | 50% | 100% | MICROSOFT DP 420 | 0% | 0% |
| SOLARWINDS NPM | 50% | 100% | MICROSOFT MB 320 | 0% | 0% |
| SPLUNK SPLK 3003 | 50% | 100% | MICROSOFT MS 201 | 0% | 0% |
| VMWARE 3V0 752 | 50% | 100% | MICROSOFT PL 200 | 0% | 0% |
| MICROSOFT DP 100 | 50% | 83% | NETSUITE NETSUITE ERP CONSULTANT | 0% | 0% |
| AMAZON AWS CERTIFIED SOLUTIONS ARCHITECT PROFESSIONAL SAP C02 | 50% | 75% | PEGASYSTEMS PEGAPCSA87V1 | 0% | 0% |
| ATLASSIAN ACP 100 | 50% | 75% | SAP C ARSOR 2202 | 0% | 0% |
| AXIS COMMUNICATIONS ANVE | 50% | 75% | SERVICENOW CIS HR | 0% | 0% |
| BLOCKCHAIN CBDH | 50% | 75% | TEST PREP GRE SECTION 2 QUANTITATIVE | 0% | 0% |
| BLUE PRISM AD01 | 50% | 75% | TEST PREP PRAXIS WRITING SECTION | 0% | 0% |
| BLUE PRISM ARA01 | 50% | 75% | UIPATH UIARD | 0% | 0% |
| CHECKPOINT 156 11577 | 50% | 75% | VMWARE 1V0 8120 | 0% | 0% |
| CIW 1D0 61B | 50% | 75% | VMWARE 2VB 602 | 0% | 0% |
| CYBERARK EPM DEF | 50% | 75% | VMWARE 3V0 2121 | 0% | 0% |
| DELL E20 624 | 50% | 75% | VMWARE 5V0 31 22 | 0% | 0% |
| DELL E20 807 | 50% | 75% | VMWARE 5V0 6119 | 0% | 0% |
| DSCI DCPLA | 50% | 75% | | | |
| DSCI DCPP 01 | 50% | 75% | | | |
| GARP SCR | 50% | 75% | | | |

| Certification | Turbo-GPT3.5 | GPT3 | Certification | Turbo-GPT3.5 | GPT3 |
|---|---|---|---|---|---|
| GENESYS GCP GC ADM | 50% | 75% | | | |
| GENESYS GCP GC ARC | 50% | 75% | | | |
| GIAC GPYC | 50% | 75% | | | |
| HP HP0 Y52 | 50% | 75% | | | |
| HP HP2 Z31 | 50% | 75% | | | |
| HP HPE0 P26 | 50% | 75% | | | |
| HP HPE6 A47 | 50% | 75% | | | |
| HP HPE6 A71 | 50% | 75% | | | |
| LPI 101 500 | 50% | 75% | | | |
| MICROSOFT 70 698 | 50% | 75% | | | |
| MICROSOFT AZ 800 | 50% | 75% | | | |
| MICROSOFT MB2 715 | 50% | 75% | | | |
| MICROSOFT MB2 717 | 50% | 75% | | | |
| MICROSOFT MS 720 | 50% | 75% | | | |
| MICROSOFT MS 900 | 50% | 75% | | | |
| SALESFORCE CRT 251 | 50% | 75% | | | |
| SERVICENOW CIS FSM | 50% | 75% | | | |
| TEST PREP ASSET | 50% | 75% | | | |
| TEST PREP PRAXIS READING SECTION | 50% | 75% | | | |
| TEST PREP TEAS SECTION 3 MATH PROBLEM SOLVING | 50% | 75% | | | |
| VERITAS VCS 273 | 50% | 75% | | | |
| VMWARE 1V0 4120 | 50% | 75% | | | |
| MICROSOFT AZ 104 | 50% | 70% | | | |